# Unsupervised domain-agnostic identification of product names in social media posts

Nicolai Pogrebnyakov
Copenhagen Business School
Frederiksberg, Denmark
nicolaip@cbs.dk

Product name recognition is a significant practical problem, spurred by the greater availability of platforms for discussing products such as social media and product review functionalities of online marketplaces. Customers, product manufacturers and online marketplaces may want to identify product names in unstructured text to extract important insights, such as sentiment, surrounding a product. Much extant research on product name identification has been domain-specific (e.g., identifying mobile phone models) and used supervised or semi-supervised methods. With massive numbers of new products released to the market every year such methods may require retraining on updated labeled data to stay relevant, and may transfer poorly across domains. This research addresses this challenge and develops a domain-agnostic, unsupervised algorithm for identifying product names based on Facebook posts. The algorithm consists of two general steps: (a) candidate product name identification using an off-the-shelf pretrained conditional random fields (CRF) model, part-of-speech tagging and a set of simple patterns; and (b) filtering of candidate names to remove spurious entries using clustering and word embeddings generated from the data.

*Keywords:* named entity recognition, social media, product names, Facebook

## 1. Introduction

Prospective customers often form opinions about a product they are contemplating to purchase or experience by reading reviews of or discussions around that product. These reviews and discussions occur in online marketplaces such as Amazon, online forums as well as social media, such as Twitter and Facebook. In many cases opinions formed by customers from these sources serve as a basis for decisions on whether they proceed with purchasing or experiencing the product.

Being able to recognize product names can play an important role in helping consumers orient themselves in the product assortment. It may also assist companies in identifying which products are popular with customers and why. Marketplaces can use this information to decide which products to carry. Automated identification of product names can be especially important in circumstances where discussion of a product is not attached to the product itself (as is the case in a marketplace, where reviews are written on product page and it is thus known which product a review belongs to), but rather contained in a stream of text, such as on forums or social media.

Accordingly, research on product name identification has been advancing. However, much of extant research has been specific to a particular domain and has relied on supervised or semi-supervised methods. At the same time, there are situations where domain specificity can be a hindrance, for instance when evaluating sentiment on products from a large online marketplace. Additionally, the rate of new product introduction is fast-paced: according to one estimate, about 30,000 new consumer products are launched each year [1]. At such rate, it can be challenging to provide new labeled training data to sustain the accuracy of product name recognition algorithms. And while in some domains or industries product names follow predictable, systematic conventions, in others that is not the case, hampering the use of rule-based identification. In some circumstances, such as chemical compounds, rule-based identification should, at first glance, be possible because of standardized nomenclatures. However, in practice they are often not followed, again precluding the use of simple heuristics in product name identification [2].

Domain-agnostic, unsupervised approaches to product name recognition may help alleviate these problems. Such approaches can be deployed in a "cold-start" fashion to analyze online reviews and social media discussions. This might be valuable in itself, as well as provide input for other algorithms, including supervised ones, that can further refine product name identification. Furthermore, being able to perform this task in an unsupervised manner helps scale information extraction [3]. However, domain-agnostic product name identification is complicated by a complete lack of convention for naming products. This holds not only across different domains, or industries, but also, as noted above, within industries and often within companies. For example, in the automotive industry most models of Ford cars are common dictionary words, such as "Ford Fusion", but there are exceptions, e.g., "Ford F-150". By contrast, car manufacturer BMW's models mostly follow a "letter digit" pattern (e.g., "BMW M4").

This study develops an unsupervised algorithm for identifying product names from social media posts on Facebook. A product is understood here as a physical product or a service that is either sold to individual customers directly (such as mobile phones), or individual customers have direct experience with it and can identify it (e.g., an aircraft model such as Boeing 747). This research formulates a small set of assumptions based on which the algorithm was developed. Under these assumptions, product name identification consists of two broad stages: (1) generate candidate product names associated with company names using simple patterns and (2) calculate several measures for each candidate name and perform clustering on these measures within each company to filter out spurious candidate names.

This study makes the following contributions:

- Develop an unsupervised, domain-agnostic algorithm for product name identification using a spectrum of unsupervised machine learning methods including word embeddings and clustering.
- Annotate a subset of product names, discuss variations in products names (which were observed even within individual companies) and evaluate the performance of the algorithm on these annotated names.

**2. Related Work**

Product name identification is an instance of a broader task of named entity recognition (NER), which aims to identify specific types of entities in a text [4]. Well-researched NER problems include recognition of proper names, names of organizations and locations [5]. Popular approaches to NER include (a) literal matching of entities within text to a dictionary, also called a gazetteer, (b) specifying a set of rules that describe an entity (e.g., "N<number>", or the letter N followed by a number, such as "N85", to match some Nokia phone models), and (c) trained models, such as conditional random fields (CRF), where a model is trained on pre-labeled data using a set of features such as individual characters within a word, capitalization etc. [2, 4, 6]. While dictionary-based approaches are easier to implement, not all entities are or can reasonably be codified in a dictionary. Without a dictionary NER becomes a non-trivial problem because of ambiguities associated with text structure, semantics and spelling [4]. These ambiguities are especially pronounced in user-generated content such as social media posts and online forum discussions, which can be particularly noisy [7, 8].

Research on identification of product names has been performed in multiple domains, ranging from consumer electronics [6, 9, 10] to programming languages and libraries [5] to chemical compound names [2]. Extant research has primarily used supervised or semi-supervised methods.

In particular, [6] used all three approaches to NER discussed above on a task that combined NER with normalization (resolving an entity to an entry in a product name catalog), and a method that combined all three achieved accuracy of 0.307. [4] used a rules-based approach to identify several categories of entities, including product names, from web pages and normalize them. Their method achieved F1 score of 0.73 across all entity categories. [5] classified software-related entities from the online community StackOverflow into five categories (such as "programming language", "API") by training a CRF model, which achieved the F1 score of 0.782. [2] used an incomplete dictionary of chemical compound names to automatically generate random names with distributions similar to those in the dictionary, then trained a CRF model with that additionally generated data. Their model achieved F1 scores of 0.882—0.945 depending on the source of data used (patents, web pages etc.).

Entity identification in social media and other user-generated content is more challenging than in formal texts such as news items, as indicated above. Existing NER CRF models pretrained on news items (e.g., Stanford NER [11]) appear to significantly leverage word capitalization, which is inconsistent in social media content such as Facebook and Twitter posts. To address this challenge, [12] combined K-nearest neighbors classifier with a CRF model and achieved F1 score of 0.725 on product name identification. [9] performed both NER and normalization of mobile phone names, and the NER portion consisted of two steps. First, a list of candidate product names was generated, which included some noise, or spurious items that were not in fact product names. At the second step a two-class CRF model was trained to differentiate true product names from the noise. That method achieved F1 score of 0.85. [3] also used a two-stage approach that first generated candidate names, followed by selection of product names based on the probability of term co-occurrences, achieving the F1 score of 0.86 on a dataset of pharmaceutical products. [7] used Freebase dictionaries to generate distributions of different entities over entity types (including products), then used topic modeling (specifically, LabeledLDA) to determine the type of an entity. Learning distributions of entities bears similarities to [2], albeit it is different in that while [2] used the learned distributions to generate additional, synthetic product names, [7] used the distributions directly in modeling. Their approach allowed classifying not only entities from the dictionaries, but also entities not encountered in the dictionaries, and achieved F1 score of 0.53 in identifying product names.

While the approaches discussed above have mostly used supervised or semi-supervised approaches (except [3]), unsupervised methods can be an important addition to the landscape of methods for product name identification [2]. They can also help advance related research on products, including attitudes of customers to various products with sentiment analysis, summarization of product reviews, identification of product features and entity normalization to name just a few. Given the proliferation of products themselves, these research areas could be helped by the ability to automatically identify product names in text.

**3. Data**

The algorithm was developed using posts from Facebook. Facebook was chosen for several reasons. First, most companies have presence on the platform, and for most consumer-facing companies such presence has turned into a necessity [13, 14]. Second, product names to be identified are mentioned both in company's own postings as well as user discussions on company pages, affording a rich environment for identifying product names. Added to this is the wide reach of Facebook, which had 1.94 billion monthly active users as of October 2018 [15]. Finally, using Facebook data complements extant research on product name identification, which tends to focus on product reviews from marketplaces such as Amazon, as well as product forums.

The dataset of posts was created in a stepwise process. Since this research focused on products that companies and consumers can discuss on their pages, 14 consumer-facing industries were identified. A list of 100 largest companies in each of these industries was collected from Dow Jones's Factiva service (http://global.factiva.com). Because many companies sell products under a brand name different from the company's name, a list of brands operated by each company was obtained from Reuters (http://www.reuters.com/sectors). After collecting Facebook pages for each brand and removing Facebook pages with no posts, 239 Facebook pages were retained in the dataset. For each page, posts both by companies (page owners) and by users on that page were collected. Only English-language posts were retained. Examples of posts are shown in Table I.

| Post source | Post text |
|---|---|
| Company | Are you an Amazon[company] Prime[product] member getting your shopping done early? |
| Company | The unique yellow of the Hyundai[company] Veloster[product] will help you stand out wherever you go. |
| User | To my great dissatisfaction I found out that the Sony[company] Z3[product] is not allowed to be leased. |
| User | I'm going to fly soon from Montreal to Frankfurt on board of one of your Airbus[company] A340-300[product]. |

Table 1. Examples of posts by companies and users, with company and product names highlighted

The resulting dataset contained 314,773 posts by companies and 1,427,178 posts by users for these companies. The distribution of the number of posts by page is shown in Fig. 1.

## 4. Problem formulation

### A. Problem formulation

Given a set $S$ of sentences in social media posts, the task is to identify in $S$ two sets, $C$ and $P$. $C = \{c_1, c_2, \ldots, c_n\}$ is a set of company names and $P = \{P_{c1}, P_{c2}, \ldots, P_{cn}\}$ is a set of sets of product names, where $P_{ci} = \{p_{1ci}, p_{2ci}, \ldots, p_{jci}\}$, $i = 1, \ldots, n$ is a set of $j$ product names for company $c_i$.

The identification of $C$ is performed with a pretrained CRF model (whose accuracy is not assessed here) and only the identification of $P$ is the focus of this research. However, given the hierarchical relationship between $C$ and $P$, the result is a two-level taxonomy with $C$ as first-order elements and $P$ as second-order elements.

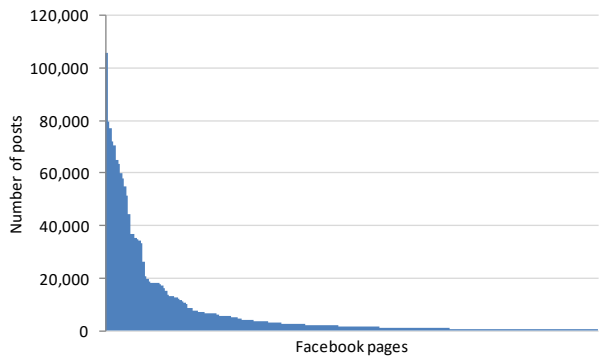

Fig. 1. Distribution of the number of posts for the Facebook pages included in the dataset (individual page names on the horizontal axis are not shown)

### B. Assumptions

The algorithm focuses only on the first word in product names $P$. Thus, if a product name contained several words, such as "ThinkPad 10", only the first word ("ThinkPad") was retained. While this might be seen as a limitation of the algorithm, many product names only contain one word (e.g., Ford's "F-150"), and for multi-word names this algorithm allows capturing names of product families that can then serve as input for identifying specific instances within these product families.

Three assumptions were formulated to facilitate the generation of candidate names from posts as explained below in the algorithm description. While these assumptions have simplified the development of the algorithm, subsequent improvements of the algorithm may focus on relaxing these assumptions.

*Assumption 1*. At least some of the mentions of product names in the social media posts will be in the form "$c_m\ p_{ncm}$", or company name followed by product name, for example: "Ford Explorer".

*Assumption 2*. The probability of occurrence of a spurious candidate name (false positive) in the pattern "$c_m\ p_{ncm}$", for example: "Microsoft's CEO", is low.

*Assumption 3*. When part-of-speech (POS) tagging is applied to posts, product names will be labeled as personal nouns in at least some of the cases.

The algorithm produces a two-level taxonomy with company names as first-order elements and product names associated with each company as second-order elements.

## 5. Product name identification algorithm

The algorithm for product name identification consists of the following stages.

I. Candidate name generation.

*1. POS tagging and company name identification with a pretrained CRF model*. Each post was tagged with a part-of-speech tagger [16] and an off-the-shelf CRF model pretrained on news articles [11]. The latter was used to identify company names. This served as input for the next step where both parts of speech and company names were used.

*2. Pattern-based candidate name identification*. In the tagged sentences, the following patterns were searched: "<Organization Name> <Proper Noun>" and "<Organization Name> <Possessive 's> <Proper Noun>". An example of the first pattern is "Microsoft Windows" and of the second, "Apple's iPhone". Whenever such patterns were encountered, both the company name (tagged with <Organization Name>) and the candidate product name (<Proper Noun>) were recorded. This resulted in a set of company names $C$ and candidate product names $P_c$ for each company $c$, and the remainder of the process focused on filtering spurious entries from the list of candidate names.

II. Filtering.

*3. Removal of misspelled and infrequent entries.* Candidate names that occurred with less than 10% frequency in each company were removed. Many such entries were revealed to be misspelled or spurious upon inspection. Additionally, companies with only one candidate product name were removed from the list, as a subsequent clustering step required at least two candidate names per company.

*4. Generation of word embeddings.* Word embeddings were created from the entire posts dataset using word2vec [17]. Each word in the dataset was lemmatized before embeddings were created, and each word was encoded as a vector of length 100. Word embeddings were used to create one of the metrics in the following step.

*5. Metrics calculation.* For each candidate product name, the following metrics were calculated: (a) cosine similarity between embedding vectors of the candidate name and the name of the company to which it was allocated; (b) inverse document frequency (IDF) score in the entire posts dataset; (c) term frequencies (TF) of occurrences of the candidate name in *other* companies. These metrics are used on the next clustering step.

*6. Clustering on the metrics.* By now the process has resulted in a list of companies, candidate product names in these companies and three metrics for each candidate name. At this step spectral clustering [18] was applied to the metrics. Two clusters were used, with the intent that one of these clusters would contain product names to be retained and the other cluster names to be excluded. Product names were first clustered on IDF, and items in the cluster with the higher average IDF score were retained. Words with higher IDF scores, which occur less frequently in the dataset, were deemed to more likely to pertain to product names than words with more frequent occurrences. The remaining names (if two or more remained) were clustered on TF, and items in the cluster with the lower average TF score were retained. This is because words that were more specific to the company under consideration, rather than to other companies, were deemed to pertain to product names. Finally, the remaining names (again, if there were two or more) were clustered on cosine similarity between the candidate product name and company name. Items in the cluster with the higher average score were retained, as it was deemed that words that occur more often in the same context as the name of the company are names of products of that company.

Fig. 2 shows the graphic representation of the algorithm.

**6. Experimental evaluation**

*A. Evaluation measures*

To assess the accuracy of the identification algorithm, posts of four companies from different industry domains were used for testing: Boeing, Ford, Lenovo and Microsoft (see Table II).

Each company had different conventions for naming products. Looking at just these four companies, it is clear that it would be difficult to create a purely rules-based product names identifier. Even for a single company there are typically variations in product naming, and these variations are substantially more pronounced across companies.

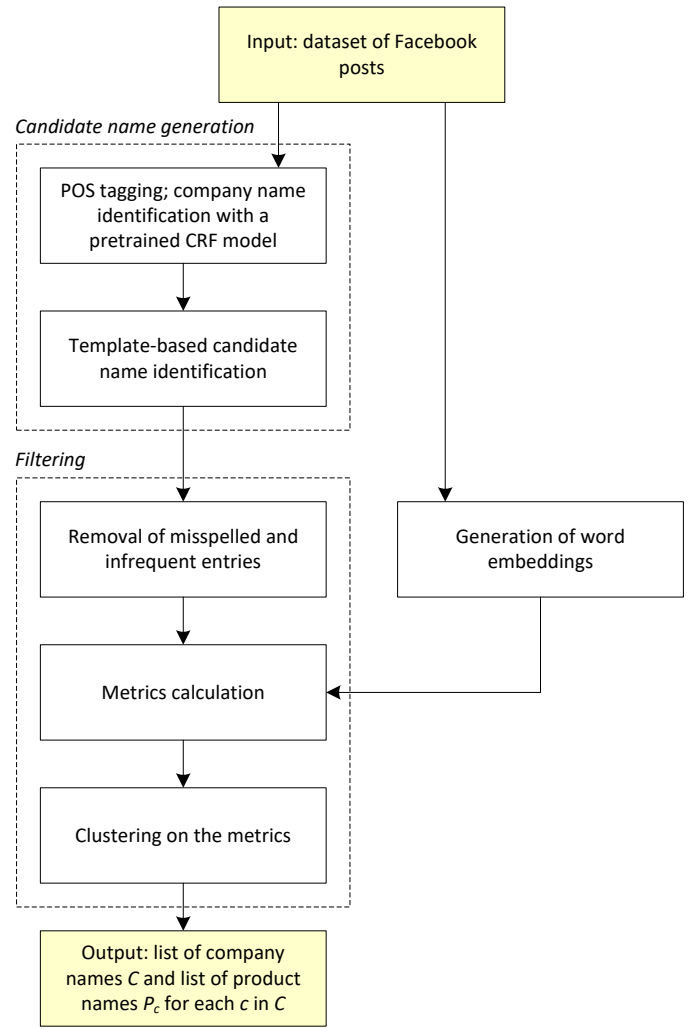

Fig. 2. Graphical representation of the algorithm

Posts from the test companies were tokenized into sentences and uploaded to Amazon's Mechanical Turk (MTurk) service. There, users were asked to identify whether a sentence contained any product names and if so, specify the product name. For example, in the sentence "The Ford Fiesta will take your family further" the users labeled "Fiesta" or "Ford Fiesta" as product name (company names were stripped from MTurk results). Names of products identified by MTurk users were then compared to names of products for these four company cases generated by the algorithm to calculate the precision, recall and F1 scores.

*B. Results*

*1) Cross-domain test cases*

The results of product name identification by each of the test companies are shown in Table III.

The results show that across all test companies, recall is higher than precision. Thus the algorithm errs on the side of caution, preferring to produce false negatives rather than misclassify an entity as a product name when in fact it is not. Indeed, in two of the four cases (Boeing and Ford) the number of false negatives is greater than true positives, and incidentally no false positives were produced. With Lenovo, the number of true positives (40) is still greater than false negatives, but the number of false negatives is significant (30), and there were a small number of false positives. In the case of Microsoft, the number of true positives is marginally greater than false negatives (17 vs. 16, respectively), while there is also a significant number of false positives (13).

| Test company | Number of true product names[a] | Example product names | Product naming convention |
|---|---|---|---|
| Boeing | 59 | (a) 737, 787<br>(b) 777-300ER, 747-8F, B-52 | (a) Numeric<br>(b) A combination of numbers, dash and letters, with numbers dominating |
| Ford | 36 | (a) Fusion, Explorer, Transit, Mustang<br>(b) F-150, C-MAX | (a) Dictionary (common) words<br>(b) Unsystematic conventions |
| Lenovo | 70 | (a) IdeaPad, Yoga<br>(b) T410, C540, X300, G470, B570e, Y580<br>(c) A7-30H | (a) Product family names. These refer not to specific products but a class of products and are typically followed by individual product name<br>(b) A letter followed by 2—3 numbers<br>(c) Unsystematic convention |
| Microsoft | 34 | (a) Windows, Office, Bing<br>(b) Xbox, HoloLens, Kinect, Cortana | (a) Dictionary (common) words<br>(b) Invented words |

[a] True column names as collected from Amazon's Mechanical Turk analysis of Facebook posts.

Table 2. Companies and their product names used for testing

*2) Effect of dataset size*

To check whether similar results could be achieved with a smaller dataset, a number of tests were run using random subsamples of the dataset of varying sizes. The subsamples ranged from 10% to 100% of the full dataset. The tests also compared the performance of different clustering methods (step 6 of the algorithm) on performance. The results are shown in Fig. 3.

The results indicate that the best performance is achieved with spectral clustering, and that dataset size matters. With spectral clustering, using larger datasets has improved the F1 score from 0.175 at 10% of the full dataset to 0.563 in the full dataset. Agglomerative clustering performed better than spectral clustering on smaller datasets, outperforming it on 10% and up to approximately 25% of the full dataset. However, the performance of agglomerative clustering reached a plateau at approximately 65% of the full dataset. K-Means, while starting at a similar performance level as spectral clustering, diverged at about 20% of the dataset size and underperformed spectral clustering, although it performed better than agglomerative clustering.

| Test company | True pos. | False neg. | False pos. | Precision | Recall | F1 |
|---|---|---|---|---|---|---|
| Boeing | 7 | 52 | 0 | 0.119 | 1 | **0.212** |
| Ford | 17 | 19 | 0 | 0.472 | 1 | **0.642** |
| Lenovo | 40 | 30 | 3 | 0.571 | 0.93 | **0.708** |
| Microsoft | 17 | 16 | 13 | 0.515 | 0.567 | **0.54** |
| Overall | 81 | 118 | 16 | 0.407 | 0.835 | **0.563** |

Table 3. Results of product name identification by test company

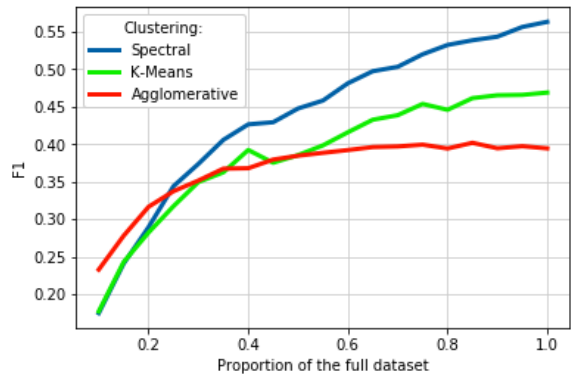

Fig. 3. Effect of dataset size and different clustering methods on algorithm performance (averages of 10 runs for each dataset size)

## 7. Discussion

This paper developed an algorithm for unsupervised identification of product names from social media posts. The algorithm consisted of two broad stages: generation of candidate product names from the corpus of social media posts and filtering the candidate names to output a list of company names, with each company name being accompanied by product names. Three assumptions were formulated to facilitate the generation of candidate product names, and it was noted that future studies may focus on relaxing these assumptions.

The candidate generation stage used an off-the-shelf CRF model pretrained on news articles to identify company names, which were used in pattern-based identification of candidate product names. Word embeddings were created, and were used at the filtering stage to calculate similarity scores between embedding vectors of each candidate product name and the corresponding company name. IDF and TF scores for each candidate product name were also calculated, and these scores were used in clustering to separate product names from spurious entries.

The algorithm exhibited an emphasis on recall rather than precision, at least in the four test cases. This could be beneficial if the algorithm is used in conjunction with other product name identification methods. For example, it can be used to generate features for other algorithms such as CRF, support vector machines (SVM), as well as provide labeled data for model training.

The balance between precision and recall was not even across the test cases, however. A possible detriment to higher precision score is that product names were not labeled as personal nouns by the POS tagger (at step 2 of the algorithm). This was particularly pronounced in the case of Boeing, which had the lowest precision score of the four test companies. Examples of product names that were not labeled as such by the algorithm include "747" or "F-15". For these products, assumption 3 did not hold. A potential future improvement for such situations is to add a pattern of "OrganizationName Number" to the patterns of candidate name identification at step 2, or add another pattern that would relax assumption 3. As for recall, the lowest score was observed for Microsoft due to a large number of false positives. This likely was a result of violated assumption 2, whereby the algorithm included into the list of candidate names many entities that were not, in fact, product names. By contrast, the high recall scores for the other three test companies suggest that assumption 1 held.

Dataset size proved to play an important role in improving the algorithm's accuracy. At first glance, this process can be applied to a dataset of any size. However, there are several benefits from having a large dataset. First, it provides a wider range of sentences, and thus a greater chance that product names will be mentioned in the context appropriate for pattern matching at step 2, and be included in candidate names. Second, it helps to train a word embeddings model that is used to determine relationships between candidate product names and company names at step 6. Finally, the performance of spectral clustering proved to increase with dataset size.

The algorithm has several advantages:

- Identified product names are linked with company name in a two-level taxonomy. For example, "Explorer" (a second-order element) is a product name of "Ford" company (first-order element).

- Since the algorithm is unsupervised and domain-agnostic, it can be extended to identification of products by companies not included in the dataset. All that is needed for this are Facebook posts discussing company products, e.g. from the company's official Facebook page or an informal "fan" Facebook page. In a similar vein, it can also be used to generate updated lists of products as they are released.

- It is not required that Facebook posts come from company-related pages (official or not) to identify products of that company. Other discussions of the company's products may be used instead (although presumably a company's products are more likely to be discussed on company-related pages).

At the same time, there are several limitations of the algorithm:

- It retains only the first word in the product name. However, for all four test companies the proportion of 1-word product names (e.g., "G460") compared to *n*-word (e.g., "Visual Studio 2010") was over 50% and ranged from 59.3% for Lenovo to 100% for Boeing (in other words, the first word uniquely identifies a product and adding extra words does not add to discriminatory power).

- It focuses only on English-language text. The ability to transfer the algorithm to other languages depends on the availability of off-the-shelf CRF models for company name tagging (which do exist, pretrained for multiple languages), as well as a sufficient number of social media posts discussing the company and its products (as dataset size was shown to affect name identification accuracy).

Future research can focus on probing deeper and addressing the above limitations, as well as extending the algorithm to perform other, related tasks such as normalization, including recognition of acronyms of product names.

REFERENCES

[1] C. Nobel. "Clay Christensen's Milkshake Marketing," October 2018; https://hbswk.hbs.edu/item/clay-christensens-milkshake-marketing.
[2] S. Yan, W. S. Spangler, and Y. Chen, "Chemical name extraction based on automatic training data generation and rich feature set," IEEE/ACM Transactions on Computational Biology and Bioinformatics, vol. 10, no. 5, pp. 1218-1233, 2013.
[3] O. Etzioni, M. Cafarella, D. Downey, A.-M. Popescu, T. Shaked et al., "Unsupervised named-entity extraction from the web: An experimental study," Artificial Intelligence, vol. 165, no. 1, pp. 91-134, 2005.
[4] R. Ananthanarayanan, V. Chenthamarakshan, P. M. Deshpande, and R. Krishnapuram, "Rule based synonyms for entity extraction from noisy text," Proceedings of the Second Workshop on Analytics for Noisy Unstructured Text Data, pp. 31-38, 2008.
[5] D. Ye, Z. Xing, C. Y. Foo, Z. Q. Ang, J. Li et al., "Software-specific named entity recognition in software engineering social content," Proceedings of the 23rd IEEE International Conference on Software Analysis, Evolution, and Reengineering (SANER), pp. 90-101, 2016.
[6] S. Wu, Z. Fang, and J. Tang, "Accurate product name recognition from user generated content," Proceedings of the 12th IEEE International Conference on Data Mining Workshops (ICDMW), pp. 874-877, 2012.
[7] A. Ritter, S. Clark, and O. Etzioni, "Named entity recognition in tweets: an experimental study," Proceedings of the Conference on Empirical Methods in Natural Language Processing, pp. 1524-1534, 2011.
[8] N. Pogrebnyakov, and E. Maldonado, "Didn't roger that: Social media message complexity and situational awareness of emergency responders," International Journal of Information Management, vol. 40, pp. 166-174, 2018.
[9] Y. Yao, and A. Sun, "Mobile phone name extraction from internet forums: a semi-supervised approach," World Wide Web, vol. 19, no. 5, pp. 783-805, 2016.
[10] S. Agrawal, K. Chakrabarti, S. Chaudhuri, and V. Ganti, "Scalable ad-hoc entity extraction from text collections," Proceedings of the VLDB Endowment, vol. 1, no. 1, pp. 945-957, 2008.
[11] J. R. Finkel, T. Grenager, and C. Manning, "Incorporating non-local information into information extraction systems by gibbs sampling," Proceedings of the 43rd Annual Meeting of the Association for Computational Linguistics, pp. 363-370, 2005.
[12] X. Liu, S. Zhang, F. Wei, and M. Zhou, "Recognizing named entities in tweets," Proceedings of the 49th Annual Meeting of the Association for Computational Linguistics, pp. 359-367, 2011.
[13] S. Iankova, I. Davies, C. Archer-Brown, B. Marder, and A. Yau, "A comparison of social media marketing between B2B, B2C and mixed business models," Industrial Marketing Management, In press.
[14] N. Pogrebnyakov, "A cost-based explanation of gradual, regional internationalization of multinationals on social networking sites," Management International Review, vol. 57, no. 1, pp. 37-64, 2017.
[15] SocialBakers. "All Facebook statistics in one place," accessed October 2018; https://www.socialbakers.com/statistics/facebook/.
[16] S. Bird, and E. Loper, "NLTK: the natural language toolkit," Proceedings of the ACL 2004 on Interactive Poster and Demonstration Sessions, p. 31, 2004.
[17] R. Rehurek, and P. Sojka, "Software framework for topic modelling with large corpora," Proceedings of the LREC 2010 Workshop on New Challenges for NLP Frameworks, 2010.
[18] A. Y. Ng, M. I. Jordan, and Y. Weiss, "On spectral clustering: Analysis and an algorithm," Proceedings of Advances in Neural Information Processing Systems, pp. 849-856, 2002.